\documentclass{article} 
\usepackage{iclr2027_conference,times}

\usepackage{amsmath,amsfonts,bm}

\def\eqref#1{equation~\ref{#1}}

\def\1{\bm{1}}

\DeclareMathAlphabet{\mathsfit}{\encodingdefault}{\sfdefault}{m}{sl}
\SetMathAlphabet{\mathsfit}{bold}{\encodingdefault}{\sfdefault}{bx}{n}

\usepackage{url}
\usepackage{booktabs}
\usepackage{amsmath,amssymb}
\usepackage{graphicx}
\usepackage{xcolor}
\usepackage{xspace}
\usepackage{multirow}
\usepackage{algorithm}
\usepackage{algpseudocode}
\usepackage[most]{tcolorbox}
\usepackage{wrapfig}
\usepackage{float}
\usepackage{needspace}
\usepackage{hyperref}
\usepackage{enumitem}

\iclrfinalcopy

\title{Critic Experience Bank: Self-Evolving Step-Level Confidence Estimation for LLM Agents}

\author{
\textbf{Yaopei Zeng}\,\textsuperscript{1}\;
\textbf{Congchao Wang}\,\textsuperscript{2}\;
\textbf{JianHang Chen}\,\textsuperscript{3}\;
\textbf{Nan Wang}\,\textsuperscript{4}\;
\textbf{Yurui Chang}\,\textsuperscript{1}\;
\textbf{Lu Lin}\,\textsuperscript{1}
\\[3pt]
{\normalfont
\textsuperscript{1}Pennsylvania State University,\quad
\textsuperscript{2}Virginia Tech,\quad
\textsuperscript{3}Purdue University,\quad
\textsuperscript{4}Amazon AGI
}
}

\newcommand{\method}{Critic Experience Bank\xspace}
\newcommand{\methodshort}{CEB\xspace}

\newcommand{\ind}{\mathbf{1}}

\begin{document}

\maketitle
\lhead{}
\renewcommand{\headrulewidth}{0pt}

\begin{abstract}
LLM agents operate in stateful environments, where a single erroneous step can waste limited interaction budget or cause irreversible effects before task failure becomes apparent.
Reliable deployment therefore requires \emph{step-level confidence estimation}: estimating, \emph{before execution}, the probability that a proposed action will advance the task.
Existing LLM confidence estimators are typically designed for static question answering under a fixed task context and evaluation criterion.
For an agent, however, its action productivity depends on an environment transition that is observed only after execution.
To address this challenge, we introduce Critic Experience Bank (CEB), a training-free framework that turns feedback from completed trajectories into reusable evidence for future confidence judgments.
After each trajectory, an LLM assigns hindsight productivity pseudo-labels to individual actions and stores them with the critic's original pre-execution confidence, task context, action, and observed feedback.
For a new action, by retrieving related productive and unproductive experiences, CEB grounds pre-execution confidence in feedback from completed trajectories to condition a fixed LLM critic.
CEB thereby adapts over a task stream without parameter updates or ground-truth step labels at deployment.
Across four agent benchmarks spanning offline and live web navigation, mobile GUI and shell tasks, and three critic backbones, CEB achieves the best or tied-best ECE, Brier score, and AUC in all twelve benchmark--backbone settings under rule-based step labels, reducing ECE by up to 53.8\% relative to the strongest training-free baseline.
Its confidence scores also improve downstream utility in selective execution and simulated task success.
\end{abstract}
\section{Introduction}
\label{sec:intro}

LLM agents increasingly solve tasks by interacting with external environments, including web interfaces, mobile GUIs, and command-line systems~\citep{deng2023mind2web,yang2023intercode}.
Unlike single-step text generation, agent actions can change the state on which subsequent decisions depend: a misplaced click can derail navigation, while submitting a form or deleting a file may have costly or irreversible consequences.
Task-level success is an insufficient reliability signal, as it reports whether the whole trajectory worked only after the interaction has unfolded.
Reliable deployment therefore calls for an earlier signal that can determine whether the next proposed action should be executed, reconsidered, or deferred to human review.

We study \emph{step-level confidence estimation}: before an action is executed, a critic estimates the probability that it will be productive for the current task. The intended notion of productivity is whether the action advances the task. 
The central challenge is that action productivity depends on an environment transition that has not yet been observed. For static question-answering, confidence assesses a completed response under a given evaluation criterion.
For an agent, confidence must instead anticipate the consequence of a proposed action, which is unavailable at scoring time and may depend on environment conditions not fully represented in the current observation.
For example, a shell command can be syntactically valid and relevant to the task yet fail because a required file or permission is missing.
The desired confidence should therefore reflect the likely productivity of the resulting interaction, rather than only the plausibility of the proposed action.
We focus on a deployment setting in which the actor and critic are fixed API models, ground-truth step labels are unavailable, and every confidence score must be produced before the proposed action is executed.

\begin{wrapfigure}{r}{0.45\linewidth}
  \centering
  \includegraphics[width=\linewidth]{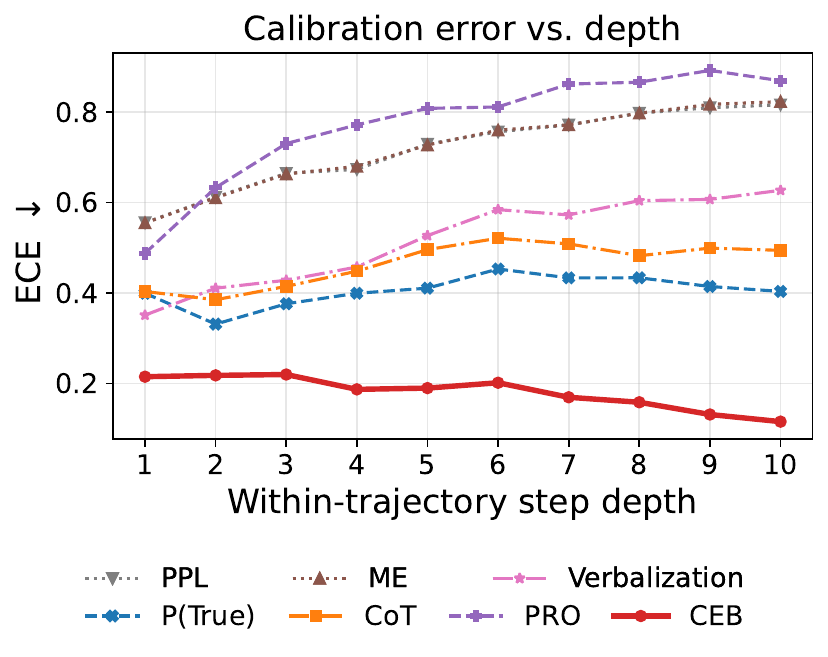}
  \vspace{-20pt}
  \caption{Step-level ECE vs.\ within-trajectory depth on Mind2Web (GPT-5.4 critic). \methodshort achieves the lowest calibration error at every depth.}
  \label{fig:depth}
  \vspace{-10pt}
\end{wrapfigure}
This dependence on environment transitions limits the direct transfer of existing LLM confidence estimators to agent actions.
Existing training-free confidence estimators derive scores from token likelihood, explicit self-evaluation, or agreement across sampled responses~\citep{kadavath2022language,tian2023just,xiong2024can}.
These approaches can be applied to agent actions, and structured reasoning can improve confidence estimation in embodied settings~\citep{yu2025uncertainty}.
However, they evaluate the current action without grounding the estimate in subsequent trajectory feedback from earlier interactions.
They may therefore remain overconfident in locally plausible actions, even when similar actions have failed to advance the task in previous trajectories.
Figure~\ref{fig:depth} illustrates this limitation on offline Mind2Web: existing confidence signals remain poorly calibrated 
across trajectory depths.

Process reward models and step-level verifiers can instead learn to assess intermediate steps through process supervision~\citep{lightman2023let,chae2025webshepherd,park2025uncertainty}.
However, this training-based approach requires labeled process data and parameter updates, which are unavailable in our setting. 
This leaves a gap between training-free but internally derived confidence signals and outcome-grounded approaches that require supervision.
The resulting question is \emph{whether a fixed critic can improve its confidence estimates by reusing feedback accumulated during deployment}.

Our key observation is that feedback from a completed trajectory arrives too late to inform its earlier actions, but can still provide evidence for future judgments.
Although the outcome of a new proposed action is not yet available, earlier interactions provide records of how related actions behaved in similar contexts.
Pairing those records with the critic's original confidence exposes discrepancies between what it predicted and what the subsequent feedback supports.
Such experience accumulated during deployment can therefore ground the future confidence estimates of a fixed critic.
When a new action resembles actions encountered in prior trajectories, the related experience can inform the estimate with evidence unavailable from the current action alone. 


We instantiate this idea as the \method (\methodshort), a training-free framework that augments a fixed LLM critic with a persistent bank of estimation-feedback records on step productivity (Figure~\ref{fig:pipeline}). Before execution, the critic assigns an initial confidence to the proposed action.
After the trajectory completes, a hindsight LLM uses the available trajectory feedback to assign productivity pseudo-labels by majority vote, which are stored together with the critic's original confidence and the corresponding task information, state, action, and feedback.
For each new action, \methodshort retrieves related productive and unproductive records to inform the critic's pre-execution estimate.
The critic thus adapts over the task stream through accumulated experience, while the critic's parameters remain fixed and no ground-truth step annotations are required at deployment.

We evaluate \methodshort on four agent benchmarks spanning offline and live web, mobile GUI, and shell tasks, using three LLM backbones.
Under rule-based step labels, \methodshort achieves the best or tied-best ECE, Brier score and AUC in all twelve benchmark--backbone combinations, reducing ECE by up to $53.8\%$ relative to the strongest training-free baseline.
These gains improve retained step accuracy and simulated task success under fixed intervention budgets.
Our contributions are as follows:
\begin{itemize}[leftmargin=*]
\item We address \emph{step-level confidence estimation} for LLM agents without ground-truth step labels at deployment: estimating a proposed action's probability of advancing the task before it is executed.
\item We introduce \methodshort, a training-free critic framework that augments a fixed LLM critic with hindsight-labeled execution experience from prior trajectories and retrieves related productive and unproductive records as precedents when future steps are scored.
\item Across four agent benchmarks and three critic backbones, \methodshort attains the best or tied-best ECE, Brier score, and AUC in all twelve dataset--critic combinations, reduces ECE by up to $54\%$ relative to the strongest training-free baseline, supports selective execution, and retains the best ranking under permissive LLM-as-judge step labels.
\end{itemize}

\section{Related Work}
\label{sec:related}

\textbf{Confidence estimation and calibration.}
Training-free LLM confidence estimation commonly uses token uncertainty, verbalized confidence, single-token probing, or probability aggregation over multiple samples~\citep{kadavath2022language,tian2023just,xiong2024can,nguyen2025probabilities}. \citet{yu2025uncertainty} adapt verbalized confidence to embodied agents with CoT prompting. These families form our baselines and can be used through APIs that expose the required outputs or token probabilities.
Token uncertainty and sample agreement are indirect signals of productivity, while explicit self-evaluation can reason about action consequences. In their standard forms evaluated here, these methods do not accumulate a persistent record linking prior confidence judgments to subsequent trajectory feedback. \methodshort retrieves such records as additional evidence for future judgments.
Calibrators such as MICE, CCPS, and DACA require hidden states or paired base and fine-tuned checkpoints~\citep{subramani2025mice,khanmohammadi2025ccps,luo2025daca}, which are unavailable in the fixed API regime. GENUINE instead learns response-level uncertainty through a supervised graph model~\citep{wang2025genuine}.
\methodshort keeps the critic fixed and instead uses execution feedback from prior trajectories as calibration evidence during inference.

\textbf{Step-level supervision and agent verification.}
Process reward models and step-level verifiers train on annotated process data~\citep{lightman2023let,chae2025webshepherd,park2025uncertainty}, and recent work studies verifier construction and statistically controlled decision rules for computer use agents~\citep{rosset2026verifiers,zhang2026glean,sadhuka2025evaluator}.
These approaches are effective when labeled process data, trainable verifiers, or calibrated validation sets exist.
We instead focus on the inference stage, where actor and critic are fixed models, ground-truth step labels are unavailable at deployment, and the score must be produced before executing the action.
\citet{mavi2025stepwise} is closest in using LLM judge scores for step-level confidence, but their judge does not build a persistent record of the critic's own interaction history.

\textbf{Experience and memory in LLM agents.}
Reflexion~\citep{shinn2023reflexion}, ExpeL~\citep{zhao2024expel}, and ReasoningBank~\citep{ouyang2026reasoningbank} accumulate agent experience to improve future action selection or reasoning, and further work studies how memory design affects agent behavior~\citep{xiong2025memory}.
These systems improve the \emph{actor}. \methodshort places memory on the \emph{critic} side, retrieving past state--action outcomes and the critic's earlier scores to estimate whether the next proposed action is reliable.
This also differs from self-refinement and tool-augmented critique~\citep{madaan2023self,bai2022constitutional,gou2023critic,wang2025steca}, which target actor improvement through critique or training rather than calibrated step-level confidence.

\section{Method}
\label{sec:method}
\begin{figure}[t]
  \centering
  \includegraphics[width=0.95\linewidth]{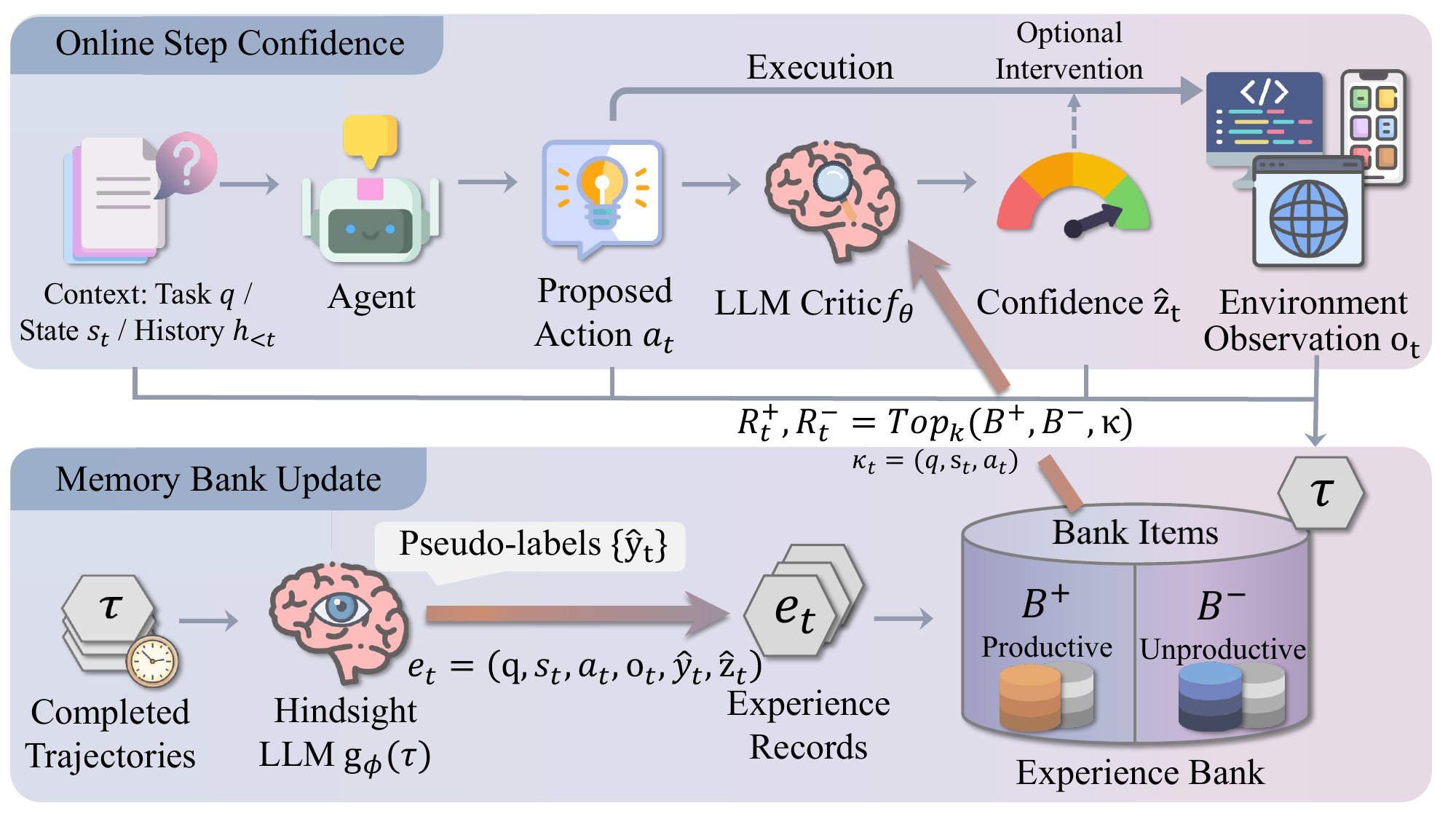}
  \caption{Overview of \methodshort.
At each step, an agent proposes an action $a_t$ based on the task, current state, and interaction history.
Before executing the action, a frozen LLM critic estimates the step confidence $\hat z_t$ using the current context and contrastive experiences retrieved from the Experience Bank.
After the full trajectory completes, a hindsight LLM labels each step as productive or unproductive from execution feedback, forming compact records that are appended to the bank.}
  \label{fig:pipeline}
\end{figure}

An action can appear reasonable without advancing the task, so confidence estimation requires evidence about its likely consequences.
Although the outcome of a proposed action is not yet available, feedback from completed trajectories can help assess related actions in similar contexts.
We therefore design \methodshort, which uses feedback from completed trajectories to inform pre-execution confidence estimates on later trajectories. It has two stages: a fixed LLM critic scores each proposed action before execution, and a hindsight LLM assesses step productivity after the trajectory ends.
The resulting assessments are stored together with the critic's original scores and retrieved to inform future step-level pre-execution confidence estimates.
Figure~\ref{fig:pipeline} summarizes this process.

\subsection{Problem Formulation}
\label{sec:method-formulation}
For a task $q$, an agent produces a trajectory
$\tau=((s_t,a_t,o_t))_{t=1}^{T}$,
where $s_t$ is the observable context before action $a_t$ and $o_t$ is the post-action feedback.
The context $s_t$ may not reveal all environmental conditions that determine an action's outcome.
Let
$h_{<t}=((s_i,a_i,o_i,\hat z_i))_{i<t}$
denote the preceding interaction history, including the confidence scores $\hat{z}$ assigned to earlier actions.

Our deployment goal is to estimate whether executing a proposed action will advance the task, before $o_{t}$ or any later trajectory information is available. For evaluation only, $y_t\in\{0,1\}$ denotes a rule-based step label used as a proxy for productivity.
Let bank $\mathcal{B}$ contain experience from previously completed trajectories. The critic produces
$\hat z_t\in[0,1]$ with the intended interpretation
\begin{equation}
\hat z_t =
f(q,s_t,h_{<t},a_t;\mathcal{B})
\;\approx\;
P(y_t=1\mid q,s_t,h_{<t},a_t,\mathcal{B}).
\label{eq:problem}
\end{equation}
The estimate can use environmental feedback only from preceding actions, but neither $o_t$ nor later information from the current trajectory. Neither $y_t$ nor any ground-truth step label is available during deployment -- we only access rule-based step labels for evaluation (\S\ref{sec:exp-setup}).

\subsection{Self-Evolving Critic via an Experience Bank}
\label{sec:method-bank}

Completed trajectories provide evidence about action productivity that was unavailable when confidence was first estimated.
\methodshort turns this evidence into a persistent memory for a fixed critic, pairing hindsight assessments with the critic's original confidence.
The hindsight assessment records whether the action was judged productive, while the original score records the critic's belief before the post-action feedback was available.
Keeping both reveals the gaps between the critic's judgement and the actual action outcome, which a productivity label alone cannot show.

\textbf{Hindsight assessments.}
Later observations and task progress can help clarify whether an earlier action was productive.
We therefore assess steps after the trajectory ends, using a hindsight LLM $g_\phi$ to interpret the completed trajectory without requiring ground-truth step annotations.
In each pass, the hindsight LLM labels every step as productive or unproductive using the available trajectory context and environment feedback.
Because the LLM may disagree across passes on ambiguous steps, we combine $V$ passes by majority vote:
\begin{equation}
\hat y_t =
\ind\Bigl[
\tfrac{1}{V}\textstyle\sum_{v=1}^{V}\hat y_t^{(v)}
\ge \tfrac12
\Bigr].
\label{eq:vote}
\end{equation}
The resulting $\hat y_t$ is a hindsight pseudo-label, distinct from the evaluation label $y_t$.

\textbf{Experience records.}
To help the critic understand and reuse a past assessment, each record stores the task, pre-action context, action, compact post-action feedback, hindsight pseudo-label, and the critic's original confidence:
\begin{equation}
e = (q',s',a',\tilde o',\hat y',\hat z'),
\label{eq:record}
\end{equation}
where $q'$ is the task, $s'$ is a compact pre-action context, $a'$ is the recorded action, and $\tilde o'$ summarizes the available post-step feedback.
The task, context, and action help retrieve similar situations, while the feedback helps the critic interpret the hindsight label.
Summarizing the context and feedback keeps the records short enough to include in the critic's prompt.
The original confidence $\hat z'$ lets the critic compare its earlier prediction with the hindsight label when judging a related action.
The retrieved record thus shows how the past action was assessed and how confident the critic had been.

\textbf{Memory updates.}
The bank starts empty and stores records from completed trajectories so that experience from one task can inform judgments on later tasks.
It remains fixed within each trajectory and receives new records only after that trajectory ends and hindsight labeling is complete.
This update schedule prevents later environment feedback from being leaked and used to score earlier actions in the same trajectory.
The critic thus adapts by using new experience records, while its parameters and prompt templates remain fixed.

\begin{algorithm}[t]
\caption{\methodshort: self-evolving critic for step-level confidence estimation.}
\label{alg:ceb}
\begin{algorithmic}[1]
\Require Task stream $\mathcal{T}$, critic $f_\theta$, hindsight LLM $g_\phi$, retrieval depth $k$, vote count $V$, bank $\mathcal{B}\leftarrow\emptyset$
\For{each task $q \in \mathcal{T}$}
  \State $\text{traj}\leftarrow\emptyset$, $\text{scores}\leftarrow\emptyset$
  \For{each step $a_t$ proposed by the agent}
    \State $s_t \gets \texttt{StateSummary}(\text{current environment or replay state})$
    \State $h_{<t}\gets$ prior entries in $\text{traj}$ paired with $\text{scores}$
    \State $\kappa_t \gets$ composite key $\bigl(\texttt{task:}~q \,\Vert\, \texttt{state:}~s_t \,\Vert\, \texttt{action:}~a_t\bigr)$
    \State $\mathcal{R}_t^{+}, \mathcal{R}_t^{-} \gets \text{TopK}_{\kappa_t}(\mathcal{B}, k)$ \Comment{$k$ per class}
    \State $\rho_t \gets \ind[a_t \text{ repeats a recent action}]$ \Comment{\S\ref{sec:method-retrieval}}
\State $\hat z_t \gets f_\theta\bigl(q, s_t, h_{<t}, a_t;\, \mathcal{R}_t^{\pm}, \rho_t\bigr)$ \Comment{Eq.~\ref{eq:ceb}}
    \State $\text{scores}.\text{append}(\hat z_t)$
    \State Obtain $o_t$ from execution or replay; $\text{traj}.\text{append}((s_t,a_t,o_t))$
  \EndFor
  \State $\{\hat y_t\}_{t=1}^{T} \gets$ majority vote of $V$ samples of $g_\phi(q, \text{traj})$ \Comment{Eq.~\ref{eq:vote}}
  \State $\mathcal{B}\gets \mathcal{B}\cup \{\mathrm{BuildRecord}(q,s_t,a_t,o_t,\hat y_t,\text{scores}[t]) : t{=}1..T\}$ \Comment{Eq.~\ref{eq:record}}
\EndFor
\end{algorithmic}
\end{algorithm}

\subsection{Experience-Grounded Confidence Estimation}
\label{sec:method-retrieval}

To find useful past experience, we match the task and context as well as the proposed action.
The same action string can serve different goals or have different effects under different environmental conditions.
We therefore construct a composite query
\[
\kappa_t =
\texttt{task:}~q
\mathbin{\Vert}
\texttt{state:}~s_t
\mathbin{\Vert}
\texttt{action:}~a_t,
\]
with bank keys constructed analogously from $(q',s',a')$.
These three fields help retrieve actions proposed for similar tasks and in similar observable states.
We fit a TF-IDF representation to the current bank and rank records by cosine similarity, which matches their wording without requiring a separate embedding model.

The critic needs examples of related actions judged productive as well as those judged unproductive, so it can consider evidence both for and against the proposed action.
We therefore partition the bank by pseudo-label and retrieve the top-$k$ records from each class, forming
$\mathcal{R}_t^{\pm}=\mathcal{R}_t^{+}\cup\mathcal{R}_t^{-}$.
Retrieving separately leaves room for evidence on both sides, and the resulting class-balanced set preserves contrastive evidence.
Each record includes the original confidence, allowing the critic to compare its earlier prediction with the later assessment.

The fixed critic combines these records with the current task, observable context, proposed action, and interaction history:
\begin{equation}
\hat z_t =
f_\theta\!\bigl(q,s_t,h_{<t},a_t;
\mathcal{R}_t^{\pm},\rho_t\bigr).
\label{eq:ceb}
\end{equation}
The history provides feedback and confidence scores from earlier steps in the current task, while retrieved records provide experience from previous tasks.
Together, they let the critic use both the current interaction and similar past situations to judge the proposed action.
We also provide a repetition flag $\rho_t$ to highlight when the proposed action repeats a recent action, helping the critic recognize possible short loops.
The critic uses this combined evidence to estimate productivity before the proposed action's outcome is available.

Algorithm~\ref{alg:ceb} summarizes the scoring and memory-update procedure.
Implementation details are provided in Appendix~\ref{app:method-details}, with web-agent prompt templates in Appendix~\ref{app:prompts}.

\section{Experiments}
\label{sec:experiments}

\subsection{Setup}
\label{sec:exp-setup}

\textbf{Datasets.}
We evaluate on four agent benchmarks: \emph{Mind2Web}~\citep{deng2023mind2web} ($1{,}009$ offline web trajectories), \emph{Mind2Web-Live}~\citep{pan2024webcanvas} ($232$ live web tasks), \emph{AMEX}~\citep{chai2024amex} ($1{,}000$ mobile-GUI episodes), and \emph{InterCode-Bash}~\citep{yang2023intercode} ($200$ shell tasks).
The offline and live settings provide complementary tests using recorded contexts and interactive execution, respectively. Dataset selection and replay protocols are detailed in Appendix~\ref{app:exp-details}.

\textbf{Models and configuration.}
We use ReAct agents~\citep{yao2023react} with GPT-4.1, GPT-5.4, or Qwen3-Next-80B-A3B-Instruct. The critic and the hindsight labeler $g_\phi$ use the same backbone as the actor.
Unless stated otherwise, \methodshort uses $V{=}5$ hindsight votes and retrieves $k{=}2$ records per class.
Additional implementation and serving details appear in Appendices~\ref{app:method-details} and~\ref{app:exp-details}. 

\textbf{Evaluation.}
We define \emph{rule-based step labels} as binary evaluation targets computed by applying predefined rules to reference annotations or environment feedback.
These rules use reference matching on Mind2Web and AMEX, key-node progress on Mind2Web-Live, and reward-based criteria on InterCode-Bash.
These labels serve as proxies for action productivity and are distinct from the hindsight pseudo-labels used to build the experience bank.
We also evaluate the same confidence scores under a separate \emph{LLM-as-judge} labeling policy on all four benchmarks.
We report Expected Calibration Error (ECE)~\citep{guo2017calibration} for calibration, Brier score for probabilistic accuracy, and AUC for ranking quality.
Label definitions and the judge protocol are given in Appendices~\ref{app:exp-details} and~\ref{sec:label-robustness}.

\textbf{Baselines and comparison protocol.}
We compare against six training-free baselines: token-based PPL and ME~\citep{heo2025uncertainty}, Verbalization~\citep{tian2023just,xiong2024can}, P(True)~\citep{kadavath2022language}, CoT (Inductive)~\citep{yu2025uncertainty}, and PRO~\citep{nguyen2025probabilities}.
All scores use only pre-execution information.
PPL, ME, P(True), CoT, and \methodshort score the same frozen actions. In the main comparisons, Verbalization and PRO generate and are evaluated on their own actions.
Baseline formulas and implementation details are in Appendix~\ref{app:baseline-details}.

\begin{table}[tbp]
\centering
\caption{Step-level confidence estimation across four benchmarks and three critic backbones under rule-based step labels. \methodshort achieves the best or tied-best ECE, Brier score, and AUC in every setting. \textbf{Bold} marks the best value per column within each critic block.}
\label{tab:main}
\small
\setlength{\tabcolsep}{4pt}
\resizebox{\textwidth}{!}{%
\begin{tabular}{ll|ccc|ccc|ccc|ccc}
\toprule
& & \multicolumn{3}{c|}{\textbf{Mind2Web}} & \multicolumn{3}{c|}{\textbf{Mind2Web-Live}} & \multicolumn{3}{c|}{\textbf{AMEX}} & \multicolumn{3}{c}{\textbf{InterCode-Bash} } \\
\textbf{Model} & \textbf{Method} & ECE$\downarrow$ & Brier$\downarrow$ & AUC$\uparrow$ & ECE$\downarrow$ & Brier$\downarrow$ & AUC$\uparrow$ & ECE$\downarrow$ & Brier$\downarrow$ & AUC$\uparrow$ & ECE$\downarrow$ & Brier$\downarrow$ & AUC$\uparrow$ \\
\midrule
\multirow{7}{*}{GPT-4.1}
 & PPL & 0.642 & 0.622 & 0.365 & 0.575 & 0.526 & 0.637 & 0.305 & 0.305 & 0.703 & 0.554 & 0.552 & 0.482 \\
 & ME & 0.646 & 0.614 & 0.364 & 0.543 & 0.487 & 0.635 & 0.301 & 0.295 & 0.711 & 0.534 & 0.532 & 0.486 \\
 & PRO & 0.342 & 0.327 & 0.581 & \textbf{0.254} & 0.261 & 0.441 & 0.608 & 0.622 & 0.495 & 0.424 & 0.424 & 0.462 \\
 & Verbalization & 0.725 & 0.669 & 0.536 & 0.682 & 0.669 & 0.481 & 0.311 & 0.307 & 0.661 & 0.450 & 0.436 & 0.623 \\
 & P(True) & 0.351 & 0.350 & 0.528 & 0.337 & 0.343 & 0.607 & 0.313 & 0.334 & 0.653 & 0.392 & 0.410 & 0.534 \\
 & CoT & 0.590 & 0.547 & 0.532 & 0.571 & 0.557 & 0.555 & 0.279 & 0.284 & 0.715 & 0.440 & 0.455 & 0.546 \\
 & \textbf{\methodshort} & \textbf{0.319} & \textbf{0.294} & \textbf{0.635} & \textbf{0.254} & \textbf{0.258} & \textbf{0.751} & \textbf{0.192} & \textbf{0.235} & \textbf{0.741} & \textbf{0.181} & \textbf{0.242} & \textbf{0.648} \\
\midrule
\multirow{7}{*}{GPT-5.4}
 & PPL & 0.707 & 0.691 & 0.340 & 0.440 & 0.508 & 0.368 & 0.233 & 0.275 & 0.683 & 0.563 & 0.569 & 0.595 \\
 & ME & 0.710 & 0.689 & 0.322 & 0.424 & 0.492 & 0.357 & 0.211 & 0.247 & 0.616 & 0.577 & 0.569 & 0.577 \\
 & PRO & 0.205 & 0.229 & 0.589 & 0.301 & 0.302 & 0.481 & 0.694 & 0.693 & 0.400 & 0.438 & 0.439 & 0.452 \\
 & Verbalization & 0.500 & 0.472 & 0.502 & 0.528 & 0.517 & 0.480 & 0.216 & 0.228 & 0.762 & 0.338 & 0.340 & 0.590 \\
 & P(True) & 0.396 & 0.394 & 0.623 & 0.416 & 0.415 & 0.608 & 0.232 & 0.240 & 0.691 & 0.402 & 0.409 & 0.566 \\
 & CoT & 0.459 & 0.404 & 0.596 & 0.412 & 0.401 & 0.608 & 0.199 & 0.223 & 0.765 & 0.339 & 0.380 & 0.524 \\
 & \textbf{\methodshort} & \textbf{0.176} & \textbf{0.206} & \textbf{0.704} & \textbf{0.171} & \textbf{0.225} & \textbf{0.764} & \textbf{0.183} & \textbf{0.216} & \textbf{0.776} & \textbf{0.254} & \textbf{0.301} & \textbf{0.635} \\
\midrule
\multirow{7}{*}{Qwen3-Next}
 & PPL & 0.812 & 0.804 & 0.367 & 0.652 & 0.625 & 0.511 & 0.378 & 0.375 & 0.746 & 0.571 & 0.567 & 0.524 \\
 & ME & 0.792 & 0.780 & 0.361 & 0.596 & 0.559 & 0.522 & 0.362 & 0.357 & 0.744 & 0.551 & 0.545 & 0.528 \\
 & PRO & 0.220 & 0.220 & 0.439 & 0.271 & 0.271 & 0.432 & 0.605 & 0.605 & 0.364 & 0.410 & 0.410 & 0.480 \\
 & Verbalization & 0.643 & 0.592 & 0.480 & 0.617 & 0.582 & 0.617 & 0.322 & 0.320 & 0.717 & 0.452 & 0.432 & 0.619 \\
 & P(True) & 0.537 & 0.507 & 0.401 & 0.487 & 0.486 & 0.410 & 0.447 & 0.464 & 0.435 & 0.400 & 0.429 & 0.468 \\
 & CoT & 0.483 & 0.460 & 0.579 & 0.387 & 0.418 & 0.606 & 0.283 & 0.306 & 0.680 & 0.455 & 0.464 & 0.560 \\
 & \textbf{\methodshort} & \textbf{0.161} & \textbf{0.169} & \textbf{0.712} & \textbf{0.196} & \textbf{0.246} & \textbf{0.755} & \textbf{0.195} & \textbf{0.234} & \textbf{0.753} & \textbf{0.215} & \textbf{0.276} & \textbf{0.641} \\
\bottomrule
\end{tabular}}
\end{table}

\subsection{Main Results}
\label{sec:main-results}

Table~\ref{tab:main} shows that \methodshort achieves the best or tied-best ECE, Brier score, and AUC in all twelve benchmark--critic settings.
Relative ECE reductions range from $6.7\%$ to $53.8\%$ in the eleven settings with a strict improvement.
These gains across web, mobile GUI, and shell tasks show that a fixed critic can benefit from past interaction experience across different environments and backbones. Compared with CoT, which reasons about the current step, \methodshort improves all three metrics in every setting.
These results suggest that feedback from past interactions provides useful evidence beyond reasoning about the current step when estimating action productivity.

\begin{table}[t]
\centering
\caption{Evaluation of the same confidence scores under permissive LLM-as-judge labels across four benchmarks and three critic backbones. \methodshort retains the best AUC in every setting. \textbf{Bold} marks the best value per column within each critic block.}
\label{tab:judge}
\small
\setlength{\tabcolsep}{4pt}
\resizebox{\textwidth}{!}{%
\begin{tabular}{ll|ccc|ccc|ccc|ccc}
\toprule
& & \multicolumn{3}{c|}{\textbf{Mind2Web}} & \multicolumn{3}{c|}{\textbf{Mind2Web-Live}} & \multicolumn{3}{c|}{\textbf{AMEX}} & \multicolumn{3}{c}{\textbf{InterCode-Bash}} \\
\textbf{Model} & \textbf{Method} & ECE$\downarrow$ & Brier$\downarrow$ & AUC$\uparrow$ & ECE$\downarrow$ & Brier$\downarrow$ & AUC$\uparrow$ & ECE$\downarrow$ & Brier$\downarrow$ & AUC$\uparrow$ & ECE$\downarrow$ & Brier$\downarrow$ & AUC$\uparrow$ \\
\midrule
\multirow{7}{*}{GPT-4.1}
 & PPL & 0.545 & 0.555 & 0.390 & 0.259 & 0.314 & 0.580 & 0.142 & 0.148 & 0.693 & 0.463 & 0.463 & 0.481 \\
 & ME & 0.545 & 0.546 & 0.394 & 0.219 & 0.293 & 0.586 & 0.123 & 0.141 & 0.705 & 0.442 & 0.446 & 0.485 \\
 & PRO & 0.348 & 0.357 & 0.541 & 0.577 & 0.579 & 0.476 & 0.141 & 0.149 & 0.671 & 0.515 & 0.515 & 0.476 \\
 & Verbalization & 0.576 & 0.555 & 0.532 & 0.360 & 0.368 & 0.566 & 0.136 & 0.144 & 0.685 & 0.366 & 0.360 & 0.680 \\
 & P(True) & 0.353 & 0.369 & 0.536 & 0.414 & 0.437 & 0.529 & 0.354 & 0.353 & 0.648 & 0.453 & 0.473 & 0.484 \\
 & CoT & 0.471 & 0.452 & 0.627 & 0.267 & 0.301 & 0.653 & 0.116 & 0.140 & 0.697 & 0.306 & 0.331 & 0.623 \\
 & \textbf{\methodshort} & \textbf{0.211} & \textbf{0.224} & \textbf{0.759} & \textbf{0.168} & \textbf{0.259} & \textbf{0.676} & \textbf{0.067} & \textbf{0.136} & \textbf{0.725} & \textbf{0.143} & \textbf{0.230} & \textbf{0.730} \\
\midrule
\multirow{7}{*}{GPT-5.4}
 & PPL & 0.589 & 0.602 & 0.317 & 0.227 & 0.263 & 0.521 & 0.159 & 0.163 & 0.691 & 0.378 & 0.403 & 0.498 \\
 & ME & 0.590 & 0.599 & 0.299 & 0.214 & 0.258 & 0.516 & 0.134 & 0.131 & 0.628 & 0.398 & 0.402 & 0.511 \\
 & PRO & 0.316 & 0.317 & 0.480 & 0.697 & 0.697 & 0.478 & 0.873 & 0.873 & 0.377 & 0.572 & 0.572 & 0.516 \\
 & Verbalization & 0.418 & 0.415 & 0.554 & \textbf{0.188} & 0.254 & 0.527 & 0.091 & 0.141 & 0.758 & 0.266 & 0.319 & 0.629 \\
 & P(True) & 0.314 & 0.318 & 0.699 & 0.370 & 0.377 & 0.608 & 0.158 & 0.162 & 0.776 & 0.361 & 0.367 & 0.639 \\
 & CoT & 0.335 & 0.319 & 0.690 & 0.287 & 0.326 & 0.627 & 0.104 & 0.139 & 0.764 & 0.230 & 0.286 & 0.638 \\
 & \textbf{\methodshort} & \textbf{0.076} & \textbf{0.159} & \textbf{0.835} & 0.199 & \textbf{0.235} & \textbf{0.667} & \textbf{0.057} & \textbf{0.108} & \textbf{0.800} & \textbf{0.225} & \textbf{0.270} & \textbf{0.694} \\
\midrule
\multirow{7}{*}{Qwen3-Next}
 & PPL & 0.734 & 0.738 & 0.257 & 0.327 & 0.341 & 0.583 & 0.229 & 0.231 & 0.784 & 0.415 & 0.418 & 0.535 \\
 & ME & 0.714 & 0.722 & 0.249 & 0.276 & 0.309 & 0.569 & 0.212 & 0.220 & 0.790 & 0.396 & 0.402 & 0.534 \\
 & PRO & 0.240 & 0.240 & 0.448 & 0.612 & 0.611 & 0.483 & 0.751 & 0.751 & 0.341 & 0.565 & 0.566 & 0.452 \\
 & Verbalization & 0.564 & 0.532 & 0.528 & 0.292 & 0.318 & 0.614 & 0.174 & 0.194 & 0.738 & 0.317 & 0.331 & 0.645 \\
 & P(True) & 0.576 & 0.550 & 0.322 & 0.389 & 0.422 & 0.460 & 0.510 & 0.503 & 0.376 & 0.463 & 0.476 & 0.450 \\
 & CoT & 0.400 & 0.412 & 0.611 & 0.246 & \textbf{0.283} & 0.632 & 0.146 & 0.182 & 0.733 & 0.311 & 0.337 & 0.563 \\
 & \textbf{\methodshort} & \textbf{0.094} & \textbf{0.143} & \textbf{0.828} & \textbf{0.244} & 0.286 & \textbf{0.677} & \textbf{0.080} & \textbf{0.148} & \textbf{0.803} & \textbf{0.157} & \textbf{0.233} & \textbf{0.719} \\
\bottomrule
\end{tabular}}
\end{table}

\textbf{LLM-as-judge labels.}
To test whether the gains depend on the rule-based step labels, we reevaluate the same scores under a separate, more permissive LLM-as-judge policy on all four benchmarks.
The judge considers the action and reasoning alongside the available reference and trajectory context, allowing exploratory and alternative valid steps to count as productive.
On Mind2Web and AMEX, agreement with the rule-based step labels is $74$--$82\%$, with most changes turning negatives into positives (Appendix~\ref{sec:label-robustness}).
Despite this change in the evaluation target, \methodshort retains the best AUC in all twelve settings and the best values on all three metrics in ten (Table~\ref{tab:judge}).
Its ranking advantage therefore extends beyond identifying exact reference matches.
The two exceptions occur on Mind2Web-Live: Verbalization has slightly lower ECE with GPT-5.4, and CoT slightly lower Brier score with Qwen3-Next.
Thus, the ranking advantage persists under both labeling policies, while the advantage in absolute probability estimates is not universal under LLM-as-judge labels.

\subsection{Component Ablation}
\label{sec:abl-component}

We ablate components on Mind2Web with GPT-5.4 to test which design choices make past experience useful (Table~\ref{tab:abl}).
All variants score identical $(\text{action},y)$ pairs, so the comparison measures changes in confidence estimation for the same decisions.
Removing the bank raises ECE from $0.176$ to $0.318$ and worsens both Brier score and AUC.
This supports the central premise that completed trajectories provide useful evidence beyond what the fixed critic obtains from the current task and interaction history.

How the bank is used also matters.
Retrieving only productive examples causes the largest ECE increase among the bank components ($+0.065$), supporting the inclusion of unproductive cases that can challenge a locally plausible action.
Random retrieval also degrades all three metrics relative to the full method, supporting selection by task, state, and action over supplying arbitrary bank examples.
Majority voting and the repetition flag provide smaller, consistent improvements, in line with their roles in stabilizing hindsight judgments and drawing attention to possible loops.
These comparisons support the design choices collectively, although they do not isolate prompt-length effects or establish the mechanism behind each gain.
Appendix~\ref{app:hyperparam} examines $V$/$k$ sensitivity. Appendix~\ref{app:retrieval} shows that the tested dense retriever does not outperform TF-IDF, supporting the lightweight retrieval choice.

\paragraph{Value of prior confidence.}
Retaining prior confidence improves calibration and slightly improves AUC under both labeling policies on Mind2Web (Table~\ref{tab:prior-confidence}), supporting its inclusion in retrieved records. The corresponding Mind2Web-Live ablation is reported in Appendix~\ref{app:additional-utility} (Table~\ref{tab:prior-confidence-live}).

\begin{table}[tbp]
\begin{minipage}[t]{0.5\linewidth}
\centering
\caption{Component ablation on Mind2Web. Removing the bank or its unproductive examples causes the largest increases in calibration error.}
\label{tab:abl}
\vspace{2pt}
\footnotesize
\setlength{\tabcolsep}{2.8pt}
\resizebox{\linewidth}{!}{%
\begin{tabular}{lccc}
\toprule
Variant & ECE$\downarrow$ & Brier$\downarrow$ & AUC$\uparrow$ \\
\midrule
\textbf{Full \methodshort} (reference) & \textbf{0.176} & \textbf{0.206} & \textbf{0.704} \\
$-$ contrastive (positives only) & 0.241 & 0.258 & 0.671 \\
$-$ TF-IDF retrieval (random from bank) & 0.218 & 0.236 & 0.682 \\
$-$ multiple sample vote ($V{=}1$) & 0.204 & 0.233 & 0.676 \\
$-$ repetition signal & 0.193 & 0.212 & 0.695 \\
Static (no bank) & 0.318 & 0.332 & 0.672 \\
\bottomrule
\end{tabular}}
\end{minipage}\hfill
\begin{minipage}[t]{0.48\linewidth}
\centering
\caption{Prior-confidence ablation on Mind2Web with GPT-5.4 under rule-based and LLM-as-judge labels. Using prior confidence improves all three metrics.}
\label{tab:prior-confidence}
\vspace{2pt}
\footnotesize
\setlength{\tabcolsep}{2.0pt}
\resizebox{\linewidth}{!}{%
\begin{tabular}{llccc}
\toprule
Evaluation labels & Variant & ECE$\downarrow$ & Brier$\downarrow$ & AUC$\uparrow$ \\
\midrule
Rule-based & Full \methodshort & \textbf{0.176} & \textbf{0.206} & \textbf{0.704} \\
 & w/o prior confidence & 0.189 & 0.223 & 0.703  \\
\midrule
LLM-as-judge & Full \methodshort & \textbf{0.076} & \textbf{0.159} & \textbf{0.835} \\
 & w/o prior confidence & 0.091 & 0.162 & 0.834 \\
\bottomrule
\end{tabular}}
\end{minipage}
\end{table}

\subsection{Self-Evolving Behavior over the Task Stream}

\label{sec:self-evolving}
\begin{wrapfigure}[15]{r}{0.4\linewidth}
\vspace{-20pt}
\centering
\includegraphics[width=\linewidth]{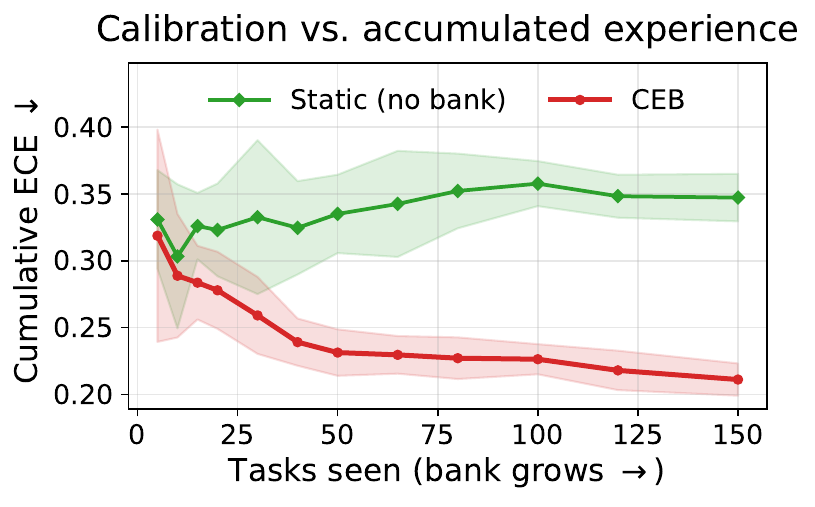}
\caption{Cumulative ECE versus the number of tasks seen on Mind2Web. \methodshort's advantage over Static grows as the bank accumulates experience.}
\label{fig:learning}
\vspace{-20pt}
\end{wrapfigure}

We test whether a fixed critic can benefit from experience accumulated during deployment, starting with an empty bank.
On Mind2Web with GPT-5.4, we compare \methodshort with the bank-free Static critic on identical action streams and track cumulative ECE over $5$ random task orders (mean$\pm$std).
As Figure~\ref{fig:learning} shows, the two begin with similar cumulative ECE, but \methodshort develops a growing advantage, reaching a gap of $0.142$ over the 150-task stream.
The increasing separation suggests that feedback from completed trajectories remains useful for later decisions, rather than benefiting only the task that produced it.
Together with the bank ablation, this supports the intended form of adaptation: the critic improves its aggregate calibration by reusing accumulated experience without parameter updates or a prefilled bank.

\subsection{Downstream Utility: Selective Execution and Task Success}
\label{sec:downstream}

Useful confidence estimates should help allocate a limited intervention budget to steps most likely to be wrong.
We first test this with a replay-based risk--coverage diagnostic on Mind2Web: defer the lowest-confidence $b\%$ of steps and measure the ground-truth positive rate among those retained.
\methodshort achieves the highest retained accuracy at every nonzero budget (Table~\ref{tab:gating}). At $b{=}50\%$, it raises accuracy from $0.194$ to $0.288$, compared with $0.234$ for CoT.
This shows that its low-confidence scores concentrate incorrect steps more effectively, making them useful targets for review or replanning.
In contrast, removing the lowest-confidence PPL or ME steps reduces retained accuracy: these signals would divert intervention toward many correct actions in this setting.
The result connects better step ranking to selective execution, while leaving the effects of actual review and subsequent environment transitions outside this replay diagnostic.
The Mind2Web-Live results further show that \methodshort's confidence remains effective for selective execution in a live environment (Appendix~\ref{app:additional-utility}).

\begin{table}[t]
\begin{minipage}[t]{0.40\linewidth}
\centering
\caption{Retained positive rate after selective execution on Mind2Web. \methodshort is best at every nonzero abstention budget.}
\label{tab:gating}
\vspace{2pt}
\footnotesize
\setlength{\tabcolsep}{3pt}
\resizebox{\linewidth}{!}{%
\begin{tabular}{lcccc}
\toprule
Method & $b{=}0$ & $10\%$ & $25\%$ & $50\%$ \\
\midrule
Perplexity & 0.194 & 0.178 & 0.154 & 0.124 \\
Mean Entropy & 0.194 & 0.175 & 0.145 & 0.114 \\
Verbalization & 0.194 & 0.198 & 0.201 & 0.219 \\
CoT & 0.194 & 0.205 & 0.222 & 0.234 \\
\textbf{\methodshort} & 0.194 & \textbf{0.212} & \textbf{0.233} & \textbf{0.288} \\
\bottomrule
\end{tabular}}
\end{minipage}\hfill
\begin{minipage}[t]{0.57\linewidth}
\centering
\caption{Task success (\%) under oracle review. \methodshort leads on Mind2Web and remains competitive on AMEX.}
\label{tab:tasksucc}
\vspace{2pt}
\footnotesize
\setlength{\tabcolsep}{3pt}
\resizebox{\linewidth}{!}{%
\begin{tabular}{lcccccccc}
\toprule
& \multicolumn{4}{c}{Mind2Web} & \multicolumn{4}{c}{AMEX} \\
\cmidrule(lr){2-5}\cmidrule(lr){6-9}
Method & $m{=}0$ & $1$ & $2$ & $3$ & $m{=}0$ & $1$ & $2$ & $3$ \\
\midrule
Perplexity & 2.1 & 4.4 & 9.5 & 20.7 & 20.6 & 28.8 & 38.2 & 49.0 \\
Mean Entropy & 2.1 & 4.2 & 9.9 & 20.9 & 20.6 & 27.0 & 32.4 & 42.0 \\
Verbalization & 2.1 & 4.9 & 11.5 & 22.8 & 20.6 & 32.0 & 43.2 & 52.6 \\
CoT & 2.1 & 4.9 & 11.6 & 22.6 & 20.6 & \textbf{33.8} & \textbf{44.7} & 53.3 \\
\textbf{\methodshort} & 2.1 & \textbf{5.5} & \textbf{12.5} & \textbf{23.7} & 20.6 & 33.4 & \textbf{44.7} & \textbf{54.6} \\
\midrule
\textit{Oracle} & 2.1 & 9.8 & 21.2 & 34.9 & 20.6 & 43.9 & 60.7 & 72.9 \\
\bottomrule
\end{tabular}}
\end{minipage}
\end{table}

\paragraph{From step-level confidence to task success.}
A useful ranking must also identify errors within each task, where one uncorrected step can prevent success.
We therefore simulate a fixed per-task review budget: an oracle reviewer corrects the $m$ lowest-confidence steps, and a task succeeds only if all its steps are then correct.
This tests whether confidence directs limited review toward the errors that matter for completing a trajectory.
\methodshort is best among confidence methods at all three budgets on Mind2Web. On AMEX, it is slightly behind CoT at $m{=}1$, ties at $m{=}2$, and leads at $m{=}3$ (Table~\ref{tab:tasksucc}).
The gains show that its ranking is useful within tasks as well as across the pooled step stream.
There is still room to improve error selection: with three reviews on Mind2Web, \methodshort reaches $23.7\%$ simulated success, compared with $34.9\%$ for the oracle selector.
Full per-method oracle-review results for Mind2Web and Mind2Web-Live are reported in Appendix~\ref{app:additional-utility} (Tables~\ref{tab:oracle-review-m2w} and~\ref{tab:oracle-review-live}). \methodshort is best or tied-best among confidence methods at every nonzero review budget.

These results assume perfect corrections on fixed trajectories and do not model the state changes that real interventions would induce.
Appendix~\ref{app:closed-loop} complements this test with actual action regeneration at fixed states: selecting steps with \methodshort recovers almost all of the oracle selector's gain at $10\%$ and $25\%$ regeneration budgets.
Together, the diagnostics support using the critic to choose where to spend review or regeneration effort.

\subsection{Closing the Gap to a Ground-Truth Oracle}
\label{sec:gt-oracle}

We test whether hindsight pseudo-labels preserve the value of experience when ground-truth step annotations are unavailable.
For this comparison only, an oracle bank replaces the hindsight labels with held-out ground-truth step labels. These annotations are never used by the main method at deployment.
We measure how much of the ECE improvement from Static to this oracle bank is recovered by \methodshort:
$(\text{Static}-\text{\methodshort})/(\text{Static}-\text{Oracle})$.
\methodshort recovers $95\%$ of the observed improvement on Mind2Web and $61\%$ on AMEX (Table~\ref{tab:gt-oracle}).
The small remaining gap on Mind2Web shows that hindsight supervision can retain most of the benefit of a ground-truth-labeled bank in this setting, supporting deployment without ground-truth step annotations.
The larger gap on AMEX shows that hindsight labels do not recover the same fraction of the benefit in every setting, leaving room to improve how experience is labeled and selected.
The oracle is an empirical reference rather than a strict bound: changing the labels also changes the class buckets and retrieved examples, so the remaining gap cannot be attributed solely to pseudo-label errors.

\begin{table}[tbp]
\centering
\caption{Ground-truth-labeled bank comparison (ECE$\downarrow$). Hindsight labels recover most of the oracle bank's calibration gain.}
\label{tab:gt-oracle}
\footnotesize
\setlength{\tabcolsep}{3.6pt}
\begin{tabular}{lcccccc}
\toprule
Dataset & Static & \textbf{\methodshort} & GT oracle & \methodshort$-$Oracle & Static$-$Oracle & Closed \\
\midrule
Mind2Web & 0.318 & \textbf{0.176} & 0.169 & 0.007 & 0.149 & $95\%$ \\
AMEX & 0.229 & \textbf{0.183} & 0.154 & 0.029 & 0.075 & $61\%$ \\
\bottomrule
\end{tabular}
\end{table}

\section{Conclusion}
\label{sec:conclusion}

We presented \methodshort, a self-evolving critic framework that recasts step-level confidence estimation for LLM agents as retrieval-augmented critic calibration over the agent's own past state--action outcomes, requiring neither parameter updates nor ground-truth step labels at deployment.
Across four datasets and three critic backbones it attains the best or tied-best ECE, Brier, and AUC in all twelve dataset--critic cells, closes up to $95\%$ of the calibration gap to a ground-truth oracle, supports selective execution and higher simulated task success under a fixed oracle-review budget, and retains the best ranking under permissive LLM-as-judge labels.
The ablation identifies the productive/unproductive contrast and state-conditioned retrieval as the primary mechanisms.


\newpage
\section*{AI use statement}
We used generative AI tools to help rephrase the manuscript and check for writing errors. We have not used generative AI tools to formulate claims or proofs, or to implement the methods. The authors verified the technical content and are responsible for it.
\bibliography{references}
\bibliographystyle{iclr2027_conference}

\newpage

\appendix


\section{LLM-as-Judge Label Analysis}
\label{sec:label-robustness}

On Mind2Web and AMEX, rule-based step labels use reference matching and mark \emph{any} action that does not match the annotated reference as unproductive, which can penalize exploratory or harmless actions and alternative valid paths too strongly.
To quantify how much this affects the evaluation target, we relabel all four benchmarks with a separate LLM-as-judge pass (GPT-5.4, temperature $0$) under a permissive policy.
For each task, the judge receives every step's agent action, the dataset reference action, and the agent's stated reasoning, batched in a single call so trajectory context is visible, and outputs one binary label per step at temperature $0$.
A step is \textsc{productive} if it serves the same intent as the reference action, constitutes reasonable harmless exploration, or makes incremental progress in the right direction. It is \textsc{unproductive} if it picks a clearly wrong control, takes a destructive or irreversible step the reference did not, or repeats a recent step without progress.

On InterCode-Bash the judge sees the task description, the benchmark's reference command (presented as one valid solution rather than the only answer), and, for every step, the agent's reasoning, the executed command, and the resulting shell output. It never sees the environment reward, the reward-derived labels, or any confidence score. A command is \textsc{productive} if it completes part of the task, gathers task-relevant information, verifies a hypothesis, or narrows the solution space, and \textsc{unproductive} if it is off-task, destructive, fails on a syntax or path error without yielding information, or repeats a previous command without new information. Alternative correct commands and necessary intermediate steps count as productive.

Table~\ref{tab:judge-shift-main} compares rule-based step labels with LLM-as-judge labels on Mind2Web and AMEX.
Across the three critics, agreement is $74$--$82\%$, and the judge mainly relabels rule-based negatives as productive.
This reflects the judge's more permissive policy. Agreement alone does not establish which labeling policy better captures true productivity.

\begin{table}[htbp]
\centering
\caption{Agreement between rule-based step labels and LLM-as-judge labels on Mind2Web and AMEX. The judge mainly relabels negatives as productive.}
\label{tab:judge-shift-main}
\small
\begin{tabular}{llccccc}
\toprule
Dataset & Agent/critic & \shortstack{Rule-based\\pos.} & \shortstack{LLM-as-judge\\pos.} & agreement & $0{\to}1$ & $1{\to}0$ \\
\midrule
Mind2Web & GPT-4.1    & 0.181 & 0.276 & $76.8\%$ & $16.4\%$ & $6.9\%$ \\
Mind2Web & GPT-5.4    & 0.194 & 0.317 & $74.5\%$ & $18.9\%$ & $6.6\%$ \\
Mind2Web & Qwen3-Next & 0.163 & 0.240 & $81.2\%$ & $13.3\%$ & $5.6\%$ \\
\midrule
AMEX     & GPT-4.1    & 0.655 & 0.840 & $73.7\%$ & $22.4\%$ & $3.9\%$ \\
AMEX     & GPT-5.4    & 0.687 & 0.875 & $81.0\%$ & $18.9\%$ & $0.2\%$ \\
AMEX     & Qwen3-Next & 0.589 & 0.751 & $82.2\%$ & $17.0\%$ & $0.8\%$ \\
\bottomrule
\end{tabular}
\end{table}

Table~\ref{tab:judge} reports all methods under LLM-as-judge labels on all four benchmarks.
\methodshort is best on all three metrics in ten of the twelve dataset--critic cells and attains the best AUC in all twelve.
The two remaining cells are on Mind2Web-Live, where the judge's permissive policy lifts the positive rate to $0.59$--$0.70$: Verbalization attains a slightly lower ECE with the GPT-5.4 critic ($0.188$ vs.\ $0.199$) and CoT a slightly lower Brier with Qwen3-Next ($0.283$ vs.\ $0.286$), while \methodshort remains best on the other two metrics in both cells.
For GPT-5.4 on Mind2Web-Live, PPL and ME include the $s{=}0.5$ imputation described in Appendix~\ref{app:exp-details}, which can affect both calibration and ranking.

\section{Method Implementation Details}
\label{app:method-details}

\paragraph{Hindsight sampling and bank updates.}
Unless stated otherwise, the hindsight labeler $g_\phi$ uses the same backbone as the critic, with $V{=}5$ samples at temperature $\eta{=}0.7$.
Each sample assigns a binary label to every step of the completed trajectory, and labels are aggregated step-wise using Eq.~\ref{eq:vote}.
For even $V$, the $\ge 1/2$ rule assigns ties to the productive class. The default odd vote count avoids ties.
The bank starts empty, remains fixed during a trajectory, and receives that trajectory's records only after completion and hindsight labeling.
Within-trajectory history can still contain feedback and confidence scores from preceding steps.
Ground-truth step labels are not used to construct the bank in the main method. The explicitly designated oracle analysis substitutes them only for comparison.

\paragraph{Record metadata.}
The implemented records augment the core fields in Eq.~\ref{eq:record} with an agreement indicator
\begin{equation}
\chi' = \ind\!\bigl[\ind[\hat z'\ge 1/2]=\hat y'\bigr].
\label{eq:agreement}
\end{equation}
This flag is displayed alongside the original score and hindsight pseudo-label to make their relationship explicit.
It records thresholded agreement, not calibration error on an individual example.
State and feedback summaries keep retrieved records compact. Prompt templates are shown in Appendix~\ref{app:prompts}.

\paragraph{Retrieval.}
The bank is partitioned as
$\mathcal{B}^{+}=\{e\in\mathcal{B}:\hat y'=1\}$ and
$\mathcal{B}^{-}=\{e\in\mathcal{B}:\hat y'=0\}$.
The retriever uses TF-IDF with 1--2 gram features fitted to the running bank's composite task--state--action keys.
The query is transformed using this representation, and records are ranked by cosine similarity.
The default $k{=}2$ is defined per class, so the prompt contains up to four records.
If a class has fewer than $k$ records, all of them are retrieved. An empty class contributes none.
An empty bank yields no retrieved examples.

\paragraph{Repetition and score output.}
The repetition flag checks for an action-prefix match between the proposed action and any of the preceding three actions, considering only available preceding steps. At the first step, $\rho_1=0$.
The critic prompt combines the task, observable context, proposed action, prior interaction history, retrieved records, and repetition cue.
Prior steps and retrieved records include their available feedback and original confidence scores.
The critic returns a JSON object with a numeric \texttt{score} field $r_t$, and the reported confidence is $\hat z_t=\mathrm{clip}(r_t,0,1)$.

\section{Experimental Details}
\label{app:exp-details}

\paragraph{Datasets and execution protocols.}
Mind2Web uses $1{,}009$ offline web trajectories with recorded page snapshots.
Mind2Web-Live instead executes actions against the live web, where subsequent page states reflect executed actions and can vary across runs.
We attempt its full $234$-task test split. Two tasks target domains whose DNS records no longer resolve for any actor, leaving $232$ tasks and $2{,}550$--$2{,}881$ steps depending on the actor.
For AMEX, we randomly sample $1{,}000$ mobile-GUI episodes. InterCode-Bash uses $200$ shell tasks.
Secondary analyses primarily use offline Mind2Web to compare variants on identical action streams, with AMEX results included where indicated.

\paragraph{Rule-based step labels.}
These binary labels define the rule-based evaluation target $y_t$ in \S\ref{sec:exp-setup}. The criteria are detailed below.
For Mind2Web-Live~\citep{pan2024webcanvas}, WebCanvas checks each trajectory against per-task \emph{key nodes} (required intermediate URLs, elements, or values). Exact key-node crediting marks only the step that reaches a node, giving a $13.7\%$ positive rate that gives no credit to preparatory steps: the steps that navigate to a node are labeled unproductive even when the node is reached. Our rule-based step labels therefore use an induced progress criterion---a step is positive iff it precedes or reaches the last key node the trajectory attains---which yields a $26$--$30\%$ positive rate across the three actors.
For the offline Mind2Web~\citep{deng2023mind2web} benchmark used in the secondary analyses, a step is positive when the agent's element prediction matches \texttt{pos\_candidates} and the operation type matches the gold action.
For AMEX~\citep{chai2024amex}, we score TAP actions and mark a step positive when the predicted touch point falls inside the human touch bounding box.
For InterCode-Bash~\citep{yang2023intercode}, the label derives from the reward function: a reward increase larger than $0.05$ is positive, and final steps with reward at least $0.7$ are positive when the reward does not decrease.
Positive rates under rule-based step labels are actor-dependent because each backbone generates different actions: they span $0.26$--$0.30$ on Mind2Web-Live, $0.16$--$0.20$ on offline Mind2Web, and $0.59$--$0.69$ on AMEX across the three actors (Table~\ref{tab:judge-shift-main}), and are $0.409$ on InterCode-Bash.

\paragraph{Baselines and scoring protocol.}
\label{app:baseline-details}
All baselines are training-free and use only pre-execution information.
The post-hoc scorers (CoT, P(True), PPL, and ME) and \methodshort score the same frozen actions against the same evaluation labels.
\methodshort processes trajectories in stream order and updates its bank only after each trajectory is completed and labeled.
For the main confidence-estimation results in Tables~\ref{tab:main} and~\ref{tab:judge}, Verbalization and PRO generate their own actions and are evaluated on those actions.

\emph{Perplexity (PPL)} and \emph{Mean Entropy (ME)}~\citep{heo2025uncertainty} aggregate token uncertainty over the action span, with confidence scores $s=1/\mathrm{PPL}$ and $s=\exp(-H_{\mathrm{mean}})$, respectively.
\emph{Verbalization}~\citep{tian2023just,xiong2024can} obtains an action and its confidence from the agent in a single forward pass, without a separate critic call.
\emph{P(True)}~\citep{kadavath2022language} asks the critic whether the proposed action is correct and reads the probability mass assigned to the \textsc{True} token.
\emph{CoT (Inductive)}~\citep{yu2025uncertainty} uses CoT reasoning without examples, followed by verbalized confidence in $[0,100]$, rescaled to $[0,1]$.
\emph{PRO}~\citep{nguyen2025probabilities} samples $N{=}5$ completions and derives confidence from a softmax over sequence log probabilities.
This adaptation differs from the original PRO entropy estimator.

\paragraph{Missing token log-probabilities on Mind2Web-Live.}
On Mind2Web-Live with the GPT-5.4 actor, the API returns an empty token-logprob payload on $1{,}289$ of $2{,}819$ planner calls ($45.7\%$, rounded to $46\%$) despite the request asking for logprobs. The omission tracks the free-form generation rather than any request parameter, and no decoding setting we tested recovers it.
PPL and ME are undefined on those steps, so we impute a neutral default $s{=}0.5$ to retain the full action stream.
The reported metrics summarize these imputed scores. Imputation can affect both calibration and ranking, so results for this setting should be interpreted with the $46\%$ missing-logprob rate in mind.
GPT-4.1 and Qwen3-Next return logprobs on every call, so nothing is imputed there.

\paragraph{Critic backbones and serving.}
GPT-4.1 and GPT-5.4 are accessed through Azure OpenAI for the offline benchmarks and through the OpenAI API for Mind2Web-Live, whose live rollouts run against BrowserBase-hosted browsers.
Qwen3-Next-80B-A3B-Instruct is served locally with vLLM on $2\times$NVIDIA H200 GPUs (tensor parallelism $2$, BF16, \texttt{gpu-memory-utilization}\,$=$\,$0.65$).

\section{Additional Ablations and Downstream Diagnostics}
\label{app:additional-utility}

\paragraph{Prior confidence on Mind2Web-Live.}
The earlier live-environment ablation remains consistent across rule-based and permissive LLM-as-judge labels: retaining prior confidence improves ECE, Brier score, and AUC under both policies (Table~\ref{tab:prior-confidence-live}).

\begin{table}[htbp]
\centering
\caption{Prior-confidence ablation on Mind2Web-Live with GPT-5.4. Retaining prior confidence improves all three metrics under both evaluation policies.}
\label{tab:prior-confidence-live}
\small
\begin{tabular}{llccc}
\toprule
Evaluation labels & Method & ECE$\downarrow$ & Brier$\downarrow$ & AUC$\uparrow$ \\
\midrule
Rule-based & Full \methodshort & \textbf{0.171} & \textbf{0.225} & \textbf{0.764} \\
 & w/o prior confidence & 0.178 & 0.234 & 0.755 \\
\midrule
LLM-as-judge & Full \methodshort & \textbf{0.199} & \textbf{0.235} & \textbf{0.667} \\
 & w/o prior confidence & 0.218 & 0.264 & 0.656 \\
\bottomrule
\end{tabular}
\end{table}

\paragraph{Selective execution on Mind2Web-Live.}
In this utility diagnostic, all methods rank the same $2{,}819$-step frozen action stream, with action correctness labels held fixed, and the lowest-confidence $b\%$ of steps are deferred. The comparison therefore measures selection on a common stream. Table~\ref{tab:selective-live} reports the positive rate. CEB achieves the best performance under different thresholds.

\begin{table}[htbp]
\centering
\caption{Retained positive rate under selective execution on Mind2Web-Live. Higher is better at a fixed deferral budget.}
\label{tab:selective-live}
\small
\begin{tabular}{lcccc}
\toprule
Method & $b{=}0$ & $b{=}10\%$ & $b{=}25\%$ & $b{=}50\%$ \\
\midrule
Perplexity & 0.303 & 0.290 & 0.265 & 0.195 \\
Mean Entropy & 0.303 & 0.290 & 0.267 & 0.189 \\
PRO & 0.303 & 0.285 & 0.270 & 0.233 \\
Verbalization & 0.303 & 0.301 & 0.302 & 0.294 \\
P(True) & 0.303 & 0.311 & 0.331 & 0.385 \\
CoT & 0.303 & 0.315 & 0.334 & 0.371 \\
\textbf{\methodshort} & 0.303 & \textbf{0.333} & \textbf{0.375} & \textbf{0.468} \\
\bottomrule
\end{tabular}
\end{table}

\paragraph{Simulated task success under oracle review.}
We simulate a fixed per-task review budget on frozen trajectories: an oracle reviewer corrects the $m$ lowest-confidence steps, without re-executing the trajectory or modeling downstream state changes. The results are therefore fixed-trajectory simulations rather than closed-loop task execution. Tables~\ref{tab:oracle-review-m2w} and~\ref{tab:oracle-review-live} give the complete results.

\begin{table}[htbp]
\begin{minipage}[t]{0.49\linewidth}
\centering
\caption{Simulated task success (\%) under oracle review on Mind2Web.}
\label{tab:oracle-review-m2w}
\vspace{2pt}
\footnotesize
\setlength{\tabcolsep}{2.4pt}
\resizebox{\linewidth}{!}{%
\begin{tabular}{lcccc}
\toprule
Method & $m{=}0$ & $m{=}1$ & $m{=}2$ & $m{=}3$ \\
\midrule
Perplexity & 2.1 & 4.4 & 9.5 & 20.7 \\
Mean Entropy & 2.1 & 4.2 & 9.9 & 20.9 \\
PRO & 2.1 & 4.4 & 9.8 & 21.1 \\
Verbalization & 2.1 & 4.9 & 11.5 & 22.8 \\
P(True) & 2.1 & 5.4 & 11.5 & 22.0 \\
CoT & 2.1 & 4.9 & 11.6 & 22.6 \\
\textbf{\methodshort} & 2.1 & \textbf{5.5} & \textbf{12.5} & \textbf{23.7} \\
\midrule
\textit{Oracle} & 2.1 & 9.8 & 21.2 & 34.9 \\
\bottomrule
\end{tabular}}
\end{minipage}\hfill
\begin{minipage}[t]{0.49\linewidth}
\centering
\caption{Simulated task success (\%) under oracle review on Mind2Web-Live.}
\label{tab:oracle-review-live}
\vspace{2pt}
\footnotesize
\setlength{\tabcolsep}{2.4pt}
\resizebox{\linewidth}{!}{%
\begin{tabular}{lcccc}
\toprule
Method & $m{=}0$ & $m{=}1$ & $m{=}2$ & $m{=}3$ \\
\midrule
Perplexity & 16.4 & 16.4 & 16.4 & 16.8 \\
Mean Entropy & 16.4 & 16.4 & 16.4 & 16.8 \\
PRO & 16.4 & 16.4 & \textbf{17.7} & 17.7 \\
Verbalization & 16.4 & \textbf{16.8} & 16.8 & 17.7 \\
P(True) & 16.4 & 16.4 & 16.4 & 16.8 \\
CoT & 16.4 & \textbf{16.8} & 16.8 & 16.8 \\
\textbf{\methodshort} & 16.4 & \textbf{16.8} & \textbf{17.7} & \textbf{18.1} \\
\midrule
\textit{Oracle} & 16.4 & 19.0 & 20.7 & 22.4 \\
\bottomrule
\end{tabular}}
\end{minipage}
\end{table}

\section{Hyperparameter Sensitivity}
\label{app:hyperparam}

We sweep the hindsight vote count $V$ and retrieval depth $k$ on Mind2Web.
At $k{=}1$, changing $V$ from $1$ to $5$ leaves ECE unchanged at the reported precision for GPT-4.1 and Qwen3-Next, and reduces it by $0.012$ for GPT-5.4 (Table~\ref{tab:vsweep}).
We retain $V{=}5$ as the default. In the component ablation at $k{=}2$, replacing it with $V{=}1$ increases ECE by $0.028$.
Since $k$ is defined per class, $k{=}1$ retrieves up to two records and $k{=}2$ up to four. These comparisons show sensitivity to voting and retrieval depth but do not isolate the mechanism behind that sensitivity.
Retrieval depth exhibits a cost--calibration tradeoff: $k{=}5$ gives the best ECE, but $k{=}2$ reaches the same ECE as $k{=}3$ with lower prompt cost (Table~\ref{tab:topk}).
We therefore deploy $k{=}2$ as the default when cost matters and treat $k{=}5$ as the setting that prioritizes calibration.
Appendix~\ref{app:token-cost} reports the corresponding token overhead.

\begin{table}[htbp]
\centering
\caption{Sensitivity to hindsight vote count. Five votes improve GPT-5.4 over a single vote, while the other critics change little.}
\label{tab:vsweep}
\footnotesize
\setlength{\tabcolsep}{2.8pt}
\begin{tabular}{l|ccc|ccc|ccc|ccc}
\toprule
 & \multicolumn{3}{c|}{$V{=}1$} & \multicolumn{3}{c|}{$V{=}2$} & \multicolumn{3}{c|}{$V{=}3$} & \multicolumn{3}{c}{$V{=}5$} \\
Critic & ECE & Br. & AUC & ECE & Br. & AUC & ECE & Br. & AUC & ECE & Br. & AUC \\
\midrule
GPT-4.1     & 0.324 & 0.294 & 0.629 & 0.327 & 0.295 & 0.627 & 0.326 & 0.295 & 0.625 & 0.324 & 0.293 & 0.630 \\
GPT-5.4     & 0.194 & 0.223 & 0.696 & 0.196 & 0.226 & 0.692 & 0.191 & 0.221 & 0.698 & 0.182 & 0.210 & 0.705 \\
Qwen3-Next  & 0.165 & 0.174 & 0.702 & 0.164 & 0.174 & 0.703 & 0.165 & 0.174 & 0.701 & 0.165 & 0.174 & 0.704 \\
\bottomrule
\end{tabular}
\end{table}

\section{Token Cost of Retrieval Depth}
\label{app:token-cost}

Each retrieved bank example contributes roughly $160$ input tokens (state summary, action, observation after action execution, past score, agreement flag), so the bank section of the prompt scales linearly in $k$ while the remainder (system prompt, current page state, history) is fixed overhead.
Table~\ref{tab:token-cost} reports measured token costs for each call on Mind2Web with GPT-5.4 (tokens measured with \texttt{tiktoken} \texttt{cl100k\_base}).
Moving from $k{=}1$ to $k{=}5$ adds $+70\%$ total input tokens for a $7\%$ relative ECE reduction.
$k{=}2$ obtains roughly half the calibration improvement at $+18\%$ tokens, which motivates it as the deployed default.

\begin{table}[htbp]
\begin{minipage}[t]{0.40\linewidth}
\centering
\caption{Sensitivity to retrieval depth. More retrieved experience lowers ECE, while AUC changes little.}
\label{tab:topk}
\footnotesize
\setlength{\tabcolsep}{4pt}
\begin{tabular}{lccc}
\toprule
$k$ & ECE$\downarrow$ & Brier$\downarrow$ & AUC$\uparrow$ \\
\midrule
$1$ & 0.182 & 0.210 & 0.705 \\
$2$ & 0.176 & 0.206 & 0.704 \\
$3$ & 0.176 & 0.204 & \textbf{0.707} \\
$5$ & \textbf{0.169} & \textbf{0.200} & 0.706 \\
\bottomrule
\end{tabular}
\end{minipage}\hfill
\begin{minipage}[t]{0.58\linewidth}
\centering
\caption{Token cost of retrieval. Increasing retrieval depth raises input cost substantially for modest calibration gains.}
\label{tab:token-cost}
\footnotesize
\setlength{\tabcolsep}{3pt}
\begin{tabular}{lcccc}
\toprule
\shortstack{$k$\\(per class)} & \shortstack{Bank\\tokens} & \shortstack{Total input\\tokens} & $\Delta$ vs.\ $k{=}1$ & ECE \\
\midrule
1 & 329   & 1{,}746 & ---      & 0.182 \\
2 & 644   & 2{,}061 & $+18\%$  & 0.176 \\
3 & 952   & 2{,}369 & $+36\%$  & 0.176 \\
5 & 1{,}554 & 2{,}971 & $+70\%$ & \textbf{0.169} \\
\bottomrule
\end{tabular}
\end{minipage}
\end{table}

\section{Reliability Diagrams}
\label{app:reliability}

\begin{figure}[htbp]
\centering
\includegraphics[width=\linewidth]{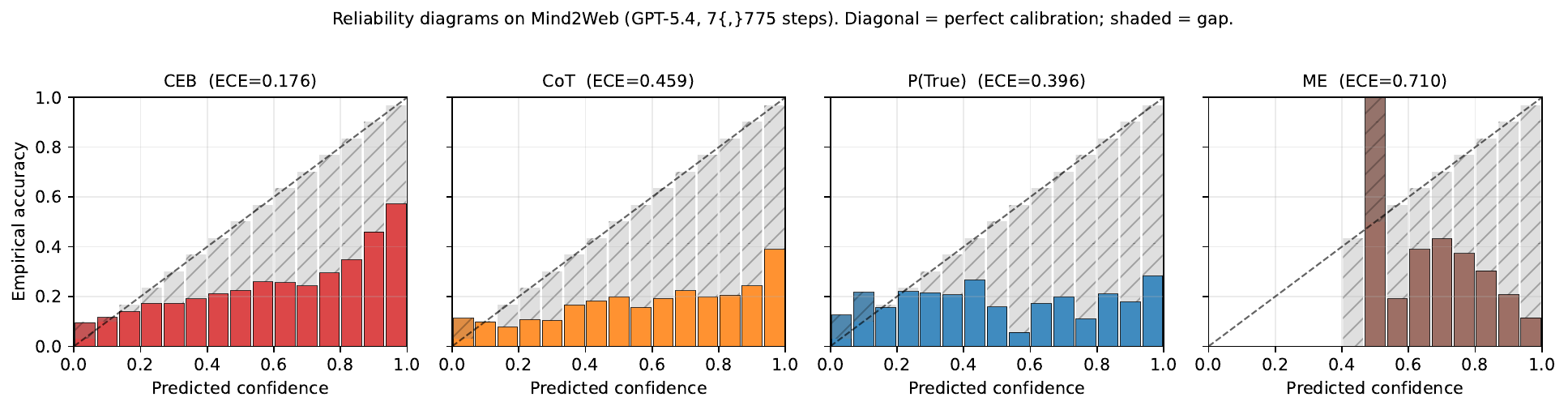}
\caption{Reliability diagrams on Mind2Web. \methodshort shows better agreement between confidence and observed step accuracy than the displayed baselines.}
\label{fig:reliability}
\end{figure}

Figure~\ref{fig:reliability} shows per-bin reliability diagrams on the full Mind2Web stream for \methodshort and three representative baselines.
\methodshort tracks the diagonal closely (ECE $0.176$), whereas CoT ($0.459$) and P(True)
($0.396$) are over-confident across most bins and Mean Entropy ($0.710$) is severely
mis-scaled, concentrating mass at high confidence regardless of correctness.
This is the bin-level view behind the aggregate ECE in Table~\ref{tab:main}.

\section{Retrieval Mechanism}
\label{app:retrieval}

\methodshort retrieves bank examples with a sparse TF-IDF score over the composite key (task $\Vert$ state $\Vert$ action).
A natural question is whether a learned dense retriever would do better.
Table~\ref{tab:retrieval} compares the deployed TF-IDF retriever against (i) a dense retriever that embeds the same composite key with OpenAI \texttt{text-embedding-3-small} and ranks by cosine similarity, and (ii) random retrieval, all on Mind2Web with the GPT-5.4 critic and an otherwise identical frozen action stream ($k{=}2$, $V{=}5$).
The dense retriever improves over random but not much, while the sparse TF-IDF key gives the best overall result.
One possible explanation is that lexical overlap preserves action strings and interface labels relevant to this task. The comparison does not establish whether this explains the gap or whether other dense retrievers would behave similarly.
TF-IDF also adds no extra model or API dependency, consistent with \methodshort's training-free design.

\begin{table}[htbp]
\centering
\caption{Retrieval comparison on Mind2Web. TF-IDF outperforms the tested dense and random retrievers.}
\label{tab:retrieval}
\small
\begin{tabular}{lccc}
\toprule
Retriever & ECE$\downarrow$ & Brier$\downarrow$ & AUC$\uparrow$ \\
\midrule
Random                                   & 0.218 & 0.236 & 0.682 \\
Dense (\texttt{text-embedding-3-small})  & 0.202 & 0.228 & 0.693 \\
\textbf{TF-IDF (deployed)}               & \textbf{0.176} & \textbf{0.206} & \textbf{0.704} \\
\bottomrule
\end{tabular}
\end{table}

\section{Offline Confidence-Gated Regeneration}
\label{app:closed-loop}

The selective-execution analysis in \S\ref{sec:downstream} tests whether low-confidence steps are good intervention targets by \emph{deferring} them.
Here we use the same selection question but test a stronger frozen-state intervention: the agent \emph{regenerates} its action on the flagged
steps, and we re-measure step correctness on the same state. On a $200$-task Mind2Web subset ($1{,}613$ steps, GPT-5.4, baseline per-step positive rate $0.183$), we re-predict each flagged step once with an identical generic
critique (``a reviewer flagged this step; reconsider''), so the regeneration procedure is held
fixed across methods and the only variable is \emph{which} steps each confidence signal selects.
We compare the five methods that score the frozen action (\methodshort, CoT, P(True), PPL, ME) at a matched regeneration budget (the lowest-confidence $b\%$ of steps), against random selection and an Oracle that prioritizes truly incorrect steps.

\begin{figure}[htbp]
\centering
\includegraphics[width=\linewidth]{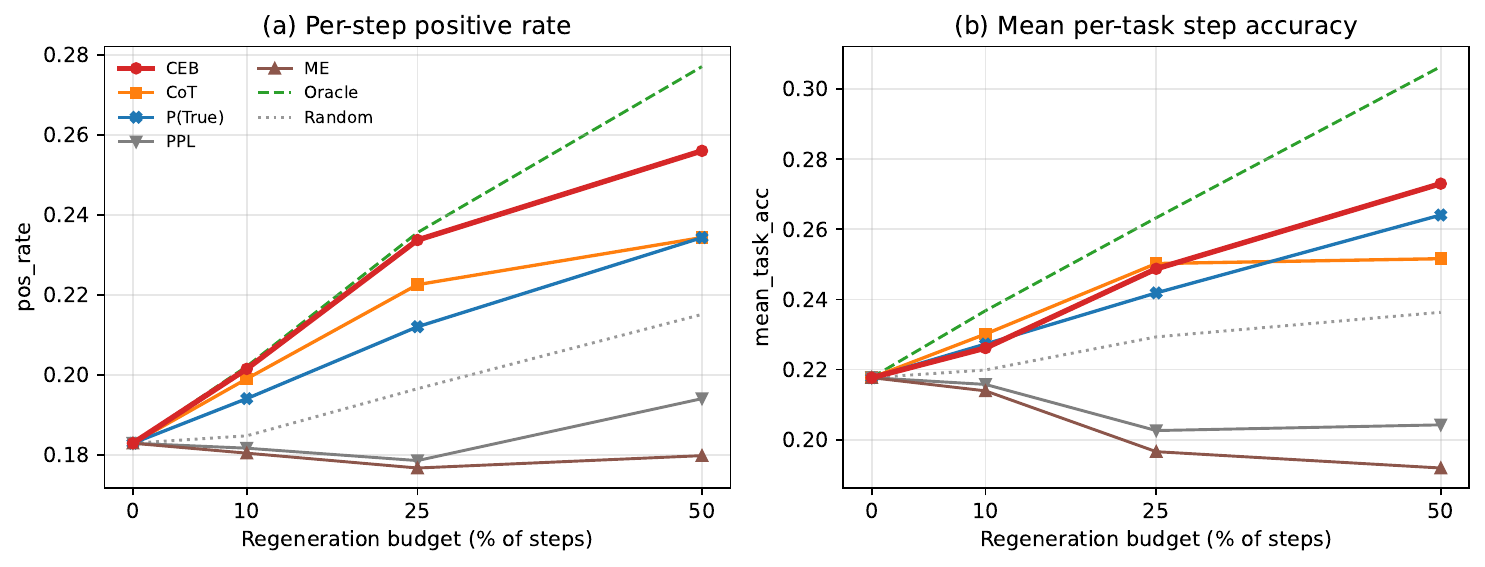}
\caption{Offline regeneration on Mind2Web: (a) step accuracy and (b) mean per-task step accuracy. \methodshort approaches the oracle at smaller regeneration budgets.}
\label{fig:closed-loop}
\end{figure}

Table~\ref{tab:closed-loop} and Figure~\ref{fig:closed-loop} report the positive rate after replacing selected steps with regenerated actions. \methodshort is the best confidence method at every budget. At $b{=}10\%$ and $25\%$, it reaches $0.201$ and $0.234$, compared with the Oracle's $0.202$ and $0.236$, recovering about $95\%$ and $96\%$ of its gain over the $0.183$ baseline. Similar aggregate gains do not imply that the two selectors choose the same steps. PPL and ME perform below random selection at all three budgets, consistent with their sub-$0.5$ AUC on offline Mind2Web.

\begin{table}[htbp]
\centering
\caption{Step accuracy after offline regeneration on Mind2Web. \methodshort selects the most effective regeneration targets among the confidence methods.}
\label{tab:closed-loop}
\small
\begin{tabular}{lccc}
\toprule
Selection & $b{=}10\%$ & $b{=}25\%$ & $b{=}50\%$ \\
\midrule
Perplexity        & 0.182 & 0.179 & 0.194 \\
Mean Entropy      & 0.180 & 0.177 & 0.180 \\
P(True)           & 0.194 & 0.212 & 0.234 \\
CoT               & 0.199 & 0.223 & 0.234 \\
\textbf{\methodshort} & \textbf{0.201} & \textbf{0.234} & \textbf{0.256} \\
\midrule
\textit{Random}  & 0.185 & 0.197 & 0.215 \\
\textit{Oracle}  & 0.202 & 0.236 & 0.277 \\
\bottomrule
\end{tabular}
\end{table}

\clearpage
\raggedbottom
\section{CEB Prompts}
\label{app:prompts}

The following figures show \methodshort's web-agent prompt templates.
Figure~\ref{fig:prompt-hindsight} shows the hindsight labeling template used after a trajectory completes. The $V{=}5$ sampled judgments produce the bank's pseudo-labels.
Figure~\ref{fig:prompt-critic} shows the pre-execution critic template, which refers to the current context, trajectory history, and retrieved contrastive examples.
The templates remain fixed during deployment. Task inputs, history, and retrieved records vary across calls.

\begin{figure}[H]
\begin{tcolorbox}[colback=gray!4, colframe=black!55,
  title={\textbf{Bank-construction prompt} (hindsight pseudo-labeler, $V{=}5$ votes per step)},
  fonttitle=\bfseries, left=2mm, right=2mm, top=1mm, bottom=1mm]
{\footnotesize
\textit{System:}
\begin{verbatim}
You are reviewing a completed web agent trajectory in hindsight.
For each step you must decide whether that step was
PRODUCTIVE (1) or UNPRODUCTIVE (0).
You have full hindsight: you can see every later step.

A step is PRODUCTIVE (1) if:
- It selected the element a careful annotator would pick
- It was a coherent step toward the task goal given the trajectory
- It was followed by progress (not retried/replaced)

A step is UNPRODUCTIVE (0) if:
- The agent had to retry or change approach
- It was a repetition that did not advance the task
- It selected an element irrelevant to the task
- The trajectory clearly went off-track

Output JSON: {"labels": [<0 or 1>, ...]}
 — exactly one integer per step, in order.
\end{verbatim}
\textit{User (template):}
\begin{verbatim}
Task: {task}

Full trajectory ({N} steps):
Step 1: [{op}] {element} | reason: {reasoning}
Step 2: ...
...

For each of the {N} steps, output 0 or 1.
JSON: {"labels": [<N values>]}
\end{verbatim}
}
\end{tcolorbox}
\caption{Hindsight prompt for assessing step productivity from completed trajectories and supplying labels for the experience bank.}
\label{fig:prompt-hindsight}
\end{figure}

\begin{figure}[H]
\begin{tcolorbox}[colback=gray!4, colframe=black!55,
  title=\textbf{Critic prompt} (pre-execution verifier with retrieved bank),
  fonttitle=\bfseries, left=2mm, right=2mm, top=1mm, bottom=1mm]
{\footnotesize
\textit{System:}
\begin{verbatim}
You are an expert evaluator of web agent actions on a webpage.

Your goal: PREDICT whether a PROPOSED next DOM action
(NOT YET executed) will advance the task. You see:
  * The task description and the CURRENT pre-action page HTML
  * Prior actions in this trajectory, each annotated with:
      - your OWN prior verifier score for that step
      - the POST-ACTION OBSERVATION (page state after execution)
  * Bank examples of similar past selections from OTHER
    trajectories. Each example shows:
      - the action that was taken
      - the STATE it was taken under (task goal + page snippet)
      - the post-action observation
      - the hindsight outcome (productive / unproductive)
      - the verifier's past score AND whether it agreed or
        disagreed with the hindsight outcome

CRITICAL — recalibrate from your own track record:
  (a) Same action in DIFFERENT states can produce different
      outcomes; weight bank examples whose STATE is closer to
      the current state more heavily.
  (b) When a bank example shows the past verifier score
      DISAGREED with the hindsight outcome, this is a known
      miscalibration pattern — adjust in the opposite direction.
  (c) If a prior in-trajectory step had a high score but its
      post-action observation shows little change, you were
      OVER-confident — adjust downward for similar selections.
  (d) Conversely, low score but clear forward progress => you
      were UNDER-confident.

Output JSON: {"score": <float 0.0-1.0>, "reason": "<one sentence>"}
Be decisive — 0.0-0.3 = unproductive, 0.4-0.6 = uncertain,
0.7-1.0 = productive.
\end{verbatim}
\textit{User (template):}
\begin{verbatim}
Task: {task}

Recent action history (with your own past scores and outcomes):
{history}

Proposed selection:
  Operation: {operation}
  Element:   {element}
  Reasoning: {reasoning}

Current page HTML (truncated):
{html}

[if repeated] NOTE: This selection looks similar to a recent
action. Repetition without progress = UNPRODUCTIVE.

Predict whether this selection will advance the task.
Output JSON: {"score": <float 0.0-1.0>, "reason": "..."}
\end{verbatim}
}
\end{tcolorbox}
\caption{Critic prompt combining the current context with past confidence scores and feedback to judge a proposed action before execution.}
\label{fig:prompt-critic}
\end{figure}

\end{document}